\documentclass[conference]{IEEEtran}
\ifCLASSINFOpdf
\else
\fi

\usepackage{amsfonts}
\usepackage{amsmath}
\usepackage{graphicx}
\usepackage{subfig}
\usepackage{xcolor}
\usepackage{newunicodechar}
\newunicodechar{ }{\,}

\begin{document}
%
\title{3D Weighted Geometric Graph Neural Networks for Sheep Facial Pain Assessment}

\author{\parbox{16cm}{\centering
    {\large Alam Noor$^{\Phi,\Psi}$, Luis Almeida$^\Psi$ and Mohamed Daoudi$^{\ddag,\P}$}\\
    {\normalsize
    $^\Phi$ CISTER Research Center, Porto, Portugal\\
    $^\Psi$ Faculty of Engineering University of Porto, Portugal\\
    $^\ddag$ Univ. Lille, CNRS, Centrale Lille, UMR 9189 CRIStAL, Lille, F-59000, France \\
    $^\P$IMT Nord Europe, Institut Mines-Télécom, Univ. Lille, Centre for Digital Systems, F-59000 Lille, France}
    }
    \thanks{This work was not supported by any organization}
}


%


\maketitle

\begin{abstract}
Deep learning systems perform mainly within the 2D for a single image domain and take the face as a single-dimension representation, losing sight of the 3D anatomy of sheep and cross-landmark spatial relationships that are intrinsic to the clinically proven Sheep Pain Facial Expression Scale (SPFES). This paper presents the \textbf{3D Sheep Pain Facial Expression System (3D-SPFES)}, a novel, monocular depth-aware geometric graph neural network system that integrates each SPFES facial landmark, such as the ears, eyes, and nose, into 3D Euclidean space estimated from a single RGB camera by using VideoDepthAnything, thus preventing the need for specialized depth hardware. Each landmark node includes a feature vector containing its 3D spatial coordinates, estimated surface normal, and facial attribute class embedding. Edges linked to nodes are assigned weights based on an aggregate metric that combines both Euclidean distance and surface co-planarity in a 3D space. A Weighted Geometric Graph Neural Network (WG-GNN) studies this graph using $\mathcal{K} = 3$ geometry-aware message-passing layers enhanced by a scaled dot-product attention method that selectively enhances anatomically relevant inter-landmark messages. The resultant node embeddings are combined into $\mathcal{O} = 3$ pain-level clusters and integrated into a Normalized Pain Score (NPS) within the range of $[0, 100\%]$ a confidence-weighted, SPFES-derived scoring method. We additionally propose bilateral detection averaging, which integrates left and right instances of symmetric face features into a singular SPFES-consistent representative node, enhancing test accuracy by 6.7\% compared to single-detection baselines. In a particular sheep facial landmark dataset, 3D-SPFES WG-GNN system model achieved a validation accuracy of 73.2\%  and an accuracy of $78.33\% \pm 2.2\%$. The proposed system can be deployed on any standard RGB camera platform, including UAVs and mobile devices, without the need for specialist sensing hardware.
\end{abstract}


%
\IEEEpeerreviewmaketitle

\section{Introduction}
\label{sec:introduction}
 
In agriculture, livestock management is important for economic productivity and ethical animal care. Sheep herds are integral across multiple cultural and farming economies globally~\cite{SMITH2024106194, NOOR2023100366}. A key strategy of effective herd management is the regular monitoring of animal health, with pain assessment recognized as a crucial indicator of welfare condition. Undiagnosed pain in sheep may give rise to systemic infection, decreased productivity, and considerable welfare loss. Therefore, early, non-invasive pain detection possesses significant clinical and societal benefits.
 
A viable approach for non-invasive pain evaluation utilizes the Sheep Pain Facial Expression Scale (SPFES)~\cite{MCLENNAN201619}, a validated clinical protocol that measures pain by assessing observable variations in 3 bilateral facial areas: ear position, eye expression, and nose shape. The SPFES encodes nine distinct facial expressions throughout these areas, each associated with a pain intensity score $p_i \in \{0, 1, 2\}$ (indicating no pain, mild pain, or severe pain). In contrast to whole-body pose estimation, the SPFES provides a direct, anatomically-based assessment of pain level that can be obtained from standardized camera image data collected by mobile devices, such manned aerial vehicles (UAVs) or static farm cameras~\cite{10155900}.
 
Previous studies on automated sheep health assessment has mainly relied on full facial detection and pose estimation employing deep learning frameworks~\cite{GUO2023108027, GU2023108143, SARWAR2021106219, LI2023107651, HITELMAN2022106713, ZHANG2022107452, 10603691}. Although these methods have encouraging detection abilities, algorithms consider the face as a singular input and fail to effectively utilize the spatial and semantic links among distinct facial landmarks. Therefore, they are unable to correctly represent the SPFES scoring approach, which requires individual pain evaluation per landmark, followed by overall assessment across landmarks.
Moreover, all present methods function within a 2D image framework, neglecting the 3D anatomical structure of the sheep's face, which is essential for distinguishing facial structures that may appear analogous in projection yet differ in spatial orientation, such as the posterior and bilateral regions of the ear or the nose bridge in relation to the jaw point.

To address these challenges, our previous work~\cite{11099391} introduced a 2D Weighted Graph Neural Network (WGNN) that represents SPFES face landmarks as graph nodes and their pain-level correlations as edges, with a cluster classification accuracy of 92.71\%. However, the proposed method is limited by mainly two challenges: \textbf{(i) it performed primarily within the 2D image plane, disregarding depth and surface spatial orientation data, and (ii) it depended on a singular camera viewpoint without any 3D geometric details, thereby reducing the predictive ability to perform edge weighting for anatomically adjacent yet structurally different facial areas.}

This paper proposes the \textbf{3D Sheep Pain Facial Expression System (3D-SPFES)}, a novel WG-GNN system model that addresses both limitations by integrating each facial landmark into 3D Euclidean space, determined by a single RGB camera utilizing a monocular depth estimation neural network. The building of a 3D graph provides geometrically accurate edge weighting, surface-normal-aware feature initialization, and equilibrium-variant message distribution. Additionally, the depth details are extracted from VideoDepthAnything~\cite{11094367}, which results in alleviating the reliance on specialized RGB-D hardware like the Intel RealSense D435. The proposed system model is easily scalable on any standardized RGB camera computing system, which includes UAVs and mobile devices.

The main scientific contribution over our previous WGNN system model~\cite{11099391} are the WG-GNN system model. Initially, we present a 3D graph representation of SPFES landmarks, wherein each node possesses spatial coordinates $\mathbf{x}_i \in \mathbb{R}^3$, surface normals $\mathbf{d}_i \in \mathbb{R}^3$, and embeddings for facial part features, while edge weights represent both metric proximity and normal alignment within 3D space. Secondly, we introduce a geometric attention method into the message-passing layers, allowing the network to precisely enhance messages from anatomically connected landmark pairs. Third, we present bilateral detection averaging, which aggregates left and right patterns of symmetrical face features into a singular representative node that conforms with the SPFES standard. Fourth, we use a 5-local-to-global model analysis method that aggregates the softmax predictions from five separately trained models, resulting in a Cohen's $\kappa$ of 0.473 (indicating strong concurrence) on the sheep facial expression dataset, which is a 163\% increase over the single-model baseline.

 
So, this work is the extension of our previous conference paper \cite{11099391}. We propose the following 4 novel minimal contributions for sheep facial expressions of pain assessment. The main contributions of this work are as follows:
 
\begin{itemize}
\item \textbf{3D facial graph representation with monocular depth.} We propose a novel 3D-SPFES graph $\mathcal{G} = (\mathcal{N}, \mathcal{E}, \mathcal{O}, \mathcal{X})$ in which each SPFES landmark node is embedded in metric 3D space using VideoDepthAnything, requiring no specialized hardware. Node features consist of 3D coordinates, projected surface normals, and embeddings of facial part categories, which provides the GNN with precise geometric structure lacking in 2D implementations.
 
\item \textbf{Geometric message passing with edge-attention weighting.} We propose a Weighted Geometric GNN (WG-GNN) layer that computes geometry-aware messages $\mathbf{m}_{ij} = \omega_{ij} \cdot \phi_m(\psi_i, \psi_j, \delta_{ij}, \mathbf{d}_i^\top\mathbf{d}_j)$, where $\omega_{ij}$ encodes both 3D proximity and surface co-planarity and $\phi_m$ is a learned edge network. A scaled dot-product attention system $\alpha_{ij}$ provides data-driven adaptation of inter-landmark message intensity levels. The analyzed attention pattern shows that the eyes node performs as the anatomical center of the face graph, corresponding with the clinical SPFES standard.
 
\item \textbf{Bilateral detection averaging for SPFES-consistent landmark representation.} We propose a systematic aggregation method that unifies all corresponding detections (left and right ear; left and right eye) per image, derives a bilateral centroid, and provides the maximum pain score from both instances. This optimizes test accuracy by 6.7\% compared to single-detection baselines.
 
\item \textbf{3D-SPFES WG-GNN evaluation with global model design.} We transform local models to a single global model with four 3D-domain augmentations, resulting in an accuracy of $78.33\% \pm 2.2\%$. A 5-local model integrated into a single model that estimates the softmax predictions of all local models obtains a test accuracy of 73.22\% with Cohen's $\kappa = 0.473$ (which means mild convergence), a mild Pain F1 score of $0.58$, and Ear $\kappa = 0.527$, representing the most substantial improvement in the process, without additional training.
 
\end{itemize}
 
The remainder of this paper is structured as follows. Section~\ref{sec:relatedwork} reviews related work on sheep facial analysis, graph neural networks for behavioural biometrics, and monocular depth estimation. Section~\ref{sec:systemmodel} presents the complete 3D-SPFES WG-GNN system model. Section~\ref{sec:experiments} describes the experimental evaluation including cross-validation, ensemble analysis, and ablation study. Section~\ref{sec:conclusion} concludes the paper and discusses directions for future work.

\section{Related Work}\label{sec:relatedwork}
In the past, researchers have studied the recognition and expression of animal faces to identify their behaviors and illnesses. These animals, particularly sheep, exhibit behavior more akin to that of humans \cite{NOOR2023100366}. These behaviors are the result of sheep's natural ability to evade potential threats and problems, as well as their human similarity, such as group living and the ability to recognize their mother and siblings, even after a long period of separation. These two factors have a significant impact on sheep's behavior and responses in a variety of situations. 

The literature studies several deep learning approaches for sheep face classification and detection. In particular, Hao et al. \cite{agriculture14030468} studied the sheep facial expression using the Single Shot MultiBox Detector (SSD) algorithm to detect the whole face of sheep and define its expressions. Ying et al. \cite{GUO2023108027} presented the detection of sheep breed using DT-YOLOv5 and were capable of recognizing facial features. 
Zishuo et al. \cite{GU2023108143} introduced a procedure that involves two phases: identification and classification. A detection network determines if each sheep's activity is normal, physiological, or disruptive, while traditional networks multi-scale feature aggregation, attention mechanism, and depthwise convolution module are used to let the network balance model size and detection accuracy. Farah et al. \cite{SARWAR2021106219} presented single as well as seven-layer convolutional neural network (CNN) models with the help of centroids by UAVs to identify the sheep. Furthermore, after fine-tuning the pre-trained models, they assembled the FCN and defined models to address recall and precision issues, while Li et al. \cite{LI2023107651} combined CNNs and vision transformers to recognise the sheep faces by extracting feature representations. Hitelman et~al. \cite{HITELMAN2022106713} studied face detection and classification using a biometric identification model. They applied faster R-CNN to detect the sheep's face in an image, while pretrained models were used to identify the face into seven distinct classification models. Zhang et~al. \cite{ZHANG2022107452} used the YOLOv4 model to identify sheep and added the convolutional block attention module (CBAM) to make the extraction of model features more stable with the help of the biometric system. In another work, Zhang et~al. \cite{10.1093/jas/skae066} used YOLOv7 with multiple attention mechanisms to detect the sheep faces. Additionally, the same authors \cite{10342186} used the YOLOv7-tiny for the multiple sheep abnormal behaviors using sheep whole body structure from a video camera. 

Similarly to the previous works, Bati et~al. \cite{Bati2024} studied the YOLOv5 model with SORT algorithm to detect the whole body of the sheep to identify and track the animal behaviors. Zhang et~al. \cite{ani13111824} also introduced YOLOv5 for the detection of the whole face of the sheep, while Ayub et~al. \cite{10469582} used YOLOv5s to classify the corresponding activity state in active and non-active for the sheep body. Zhang et~al. \cite{ZHANG2024108697} also introduced the data called multi-view sheep face images and applied the vision transform model to recognize the sheep faces with a heat map. Kelly et~al. \cite{KELLY2024110027} presented a sheep activity dataset to analyze the behaviors of the sheep with deep learning pre-trained models. Xue et~al. \cite{ani14131923} introduced a sheep face orientation recognition algorithm to different face orientations and a feature point-matching and reconstructing the sheep face. However, the algorithm requires images from different angles to accurately recreate the sheep face. Pang et~al. \cite{Pang2023} studied the feature extraction of sheep faces using an attention residual module to aggregate the features received from the input of a spherical camera to capture image data. However, the model relies on classification, making it challenging to identify sheep faces for its application. Cai et~al. \cite{10.1145/3650400.3650652} used convolutional neural networks for the sheep face recognition for disease prevention, while Xinyu et~al. \cite{10505584} presented the dilated convolutional attention module for gender identification using sheep faces as a binary classification model.

The existing models, as illustrated in the literature, focus primarily on the sheep's entire face or body. With such an approach, classifying the sheep's facial pain is very difficult. To classify the pain, we must examine every part of the sheep's entire face. Furthermore, the detection models only detect facial parts expression, requiring the clustering of painful parts and their separation from non-painful ones to define and combine the total face pain. Therefore, our study follows the novel sheep facial landmark dataset, proposing the use of WGNN to determine the total pain of the sheep faces.

\section{System Model}\label{sec:systemmodel}

\subsection{Graph Formulation and 3D Spatial Representation}

The proposed 3D-SPFES WG-GNN system model improves upon 2D facial network analysis by integrating each facial part expression (FPE) node into 3D Euclidean space, effectively and precisely representing the anatomical geometry of the sheep face. Depth-aware perception, derived from a VideoDepthAnything vision reconstruction, assigns a spatial coordinate triplet to each identified face landmark. The comprehensive 3D-SPFES graph is shown as in (\ref{eqn:graph}),
\begin{equation}\label{eqn:graph}
    \mathcal{G} = (\mathcal{N},\ \mathcal{E},\ \mathcal{O},\ \mathcal{X}),
\end{equation}
where $\mathcal{N}$ represents the set of nodes containing 9 facial expressions (eyes, ears, nose, cheeks, lips, jaw), $\mathcal{E}$ represent the set of directed edges that encode spatial and pain-level relationships across the nodes, $\mathcal{O}$ represents the set of output state labels corresponding to pain clusters; and $\mathcal{X} \in \mathbb{R}^{|\mathcal{N}| \times 3}$ be the coordinate matrix where the $i$-th row $\mathbf{x}_i = (x_i, y_i, z_i)^\top$ denotes the 3D centroid of a facial part $i$ within the camera frame. The depth dimension $z_i$ is essential: it clarifies facial parts that occupy equivalent 2D image orientations (e.g., medial cortex versus lateral cortex of the eye) and offers a more robust structural prior for graph edge generation than plain pixel proximity.

A prerecorded diagnosis label is utilized on the RGB channel to obtain per-part pain intensity labels $p_i \in \{0, 1, 2\}$ for all 9 face parts $\mathcal{P} = [p_1, p_2, \ldots, p_9]$, where 0 represents no pain, 1 denotes mild pain, and 2 reflects severe pain. A 3D back-projection process is translated from each identified bounding box centroid from image coordinates to the 3D coordinate system utilizing the depth map of VideoDepthAnything and the camera's intrinsic matrix $\mathbf{K}$ as shown in~(\ref{eqn:3d_projection}),
\begin{equation} \label{eqn:3d_projection}
    \mathbf{x}_i = z_i \cdot \mathbf{K}^{-1} \begin{bmatrix} u_i \\ v_i \\ 1 \end{bmatrix},
\end{equation}
where $(u_i, v_i)$ represents the pixel centroid of the $i$-th face part and $z_i$ represents the equivalent depth value obtained from the aligned depth frame. The surface normal $\mathbf{d}_i \in \mathbb{R}^3$ at each facial category, derived from the local point cluster proximity by principal component analysis (PCA) applied to the $k$-nearest depth points, represents the direction of the facial surface in that part. Parts on the flat nose bridge will have normals closely aligned with the camera's visual orientation, whereas ear tips will display laterally oriented normals; this geometric differentiation is utilized in edge weighting.

\subsection{3D Node Feature Initialization}

Each node $n \in \mathcal{N}$ contains a feature vector $\psi$ that performs as the input value to the geometric graph neural network. In the first 2D variant, this feature vector is limited to the scaling pain level and a categorical representation of facial-part types. The 3D node extension enhances this representation using the node's 3D spatial coordinates and surface normal, effectively connecting the graph with explicit geometric contextual information. The in-depth feature vector for face expression part $i$ is shown as~(\ref{eqn:psi_i}),
\begin{equation} \label{eqn:psi_i}
    \psi_i^{(0)} = \left[\ p_i,\ \text{FPT}_i,\ \mathbf{x}_i^\top,\ \mathbf{d}_i^\top\ \right] \in \mathbb{R}^{1+C+3+3},
\end{equation}
where $p_i \in \{0,1,2\}$ represent the pain intensity, $\text{FPT}_i \in \mathbb{R}^C$ show a one-hot or learned embedding of the facial part type (e.g., left eye, right ear, nose tip) with $C$ categories, $\mathbf{x}_i = (x_i, y_i, z_i)^\top \in \mathbb{R}^3$ represents the 3D spatial position, and $\mathbf{d}_i = (d_{ix}, d_{iy}, d_{iz})^\top \in \mathbb{R}^3$ indicates the unit surface normal vector. The complete initial feature dimensionality corresponds to $1 + C + 6$.

Adding $\mathbf{x}_i$ directly into the feature vector, instead of exclusively applying it for edge construction, allows the graph to distinguish position-dependent pain correlations. For instance, a pain score of 2 at the face might exhibit significant clinical repercussions with respect to the equivalent score at the orbital region, with the spatial coordinate providing a distinguishing parameter for the learned transformation matrices. The surface normal $\mathbf{d}_i$ adds nodes with proximate 3D coordinates but divergent anatomical orientations, such as the inner and outer jaw borders.

\subsection{3D Edge Construction with Geometric Weighting}

In the early 2D WGNN system model~\cite{11099391}, edges are assigned according to facial morphology proximity and pain-score symmetry. The 3D WG-GNN system combines the qualitative proximity criterion with a quantitative Euclidean distance threshold and incorporates a normalization factor into the edge weight, resulting in a geometrically accurate connectivity structure. An edge $e = (n, w) \in \mathcal{E}$ forms between nodes $n$ and $w$ if and only if both of the conditions $\delta_{nw} = \|\mathbf{x}_n - \mathbf{x}_w\|_2 \leq \tau$ and $|p_n - p_w| \leq \epsilon$ are satisfied simultaneously. While, the initial state requires anatomical proximity in 3D space with a threshold $\tau$, providing that only physically neighboring facial parts are considered as structurally correlated. The second prerequisite specifies pain-level regularity with tolerance $\epsilon \in \{0, 1, 2\}$, categorizing facial parts showing an analogous degree of pain. Both conditions must be at the same time observed: two anatomically distinct areas (e.g., left ear and right ear) are unlikely to be linked despite identical pain scores, and two adjacent areas with considerably different pain scores will not be connected even if they are in close proximity. After the construction of an edge, its scaling weight $\omega_{nw}$ assesses the degree of strength of the relationship by integrating a radial basis function (RBF) of 3D space with a normal-alignment attribute, as shown in~(\ref{eqn:omega_{nw}}),

\begin{equation}\label{eqn:omega_{nw}}
    \omega_{nw} = \underbrace{\exp\!\left(-\frac{\delta_{nw}^2}{2\sigma^2}\right)}_{\text{proximity term}} \cdot \underbrace{\frac{\mathbf{d}_n^\top \mathbf{d}_w + 1}{2}}_{\text{normal alignment term}}.
\end{equation}

The proximity factor is a Gaussian kernel centered at zero distance, constrained by bandwidth $\sigma$: edges between precisely positioned nodes (small $\delta_{nw}$) acquire weights equivalent to 1.0, but edges adjacent to the threshold $\tau$ are attributed weights that gradually decrease toward 0. The norm alignment metric is the normalized cosine similarity of surface normals, transformed from $[-1, +1]$ to $[0, 1]$: co-planar facial parts (parallel normals, $\mathbf{d}_n^\top \mathbf{d}_w \approx 1$) attain correlation scores close to orthogonally oriented facial parts receive scores near 0.5, and anti-parallel facial parts (physically infeasible for a convex surface but incorporating a for robustness) receive scores near 0. The multiplicative coupling of these two factors shows that a structurally key edge implies both proximity and co-planarity, which is effective in preventing defective connections across anatomical edges, such as the transition from jaw to ear.

\subsection{Optimal 3D Parse Graph Inference}

For each facial part type (FPT), a parse graph $g = (\mathcal{N}_g, \mathcal{E}_g, \mathcal{O}_g)$ is inferred as a sub-graph of $\mathcal{G}$ such that $\mathcal{N}_g \subseteq \mathcal{N}$ and $\mathcal{E}_g \subseteq \mathcal{E}$. The 3D formulation augments the inference objective with the coordinate matrix $\mathcal{X}$, so that the optimal parse graph $g^*$ is defined as shown in~(\ref{eqn:optimal_parse_graph}),

\begin{equation}\label{eqn:optimal_parse_graph}
    g^* = \arg\max_{g}\ \underbrace{P_d\!\left(\mathcal{O}_g \mid \mathcal{N}_g,\ \mathcal{E}_g,\ \psi,\ \mathcal{X}\right)}_{\text{labelling probability}} \cdot \underbrace{P_d\!\left(\mathcal{N}_g,\ \mathcal{E}_g \mid \psi,\ \mathcal{X},\ \mathcal{G}\right)}_{\text{structural probability}}
\end{equation}
where, $P_d(\mathcal{O}_g \mid \mathcal{N}_g, \mathcal{E}_g, \psi, \mathcal{X})$, is the labeling probability: given the graph structure and all node/edge features, including 3D positions.
While, $P_d(\mathcal{N}_g, \mathcal{E}_g \mid \psi, \mathcal{X}, \mathcal{G})$ represents the structural probability, assuming the feature-augmented 3D graph $\mathcal{G}$, what is the likelihood of this specific sub-graph $g$ serving as an explanation for the observed facial expression? By conditioning both factors on $\mathcal{X}$, the inference simultaneously assesses the degree to which the proposed parse graph is geometrically consistent with the sheep's true facial anatomy. For instance, parse graphs that attribute the same pain cluster to anatomically disparate facial parts exclusively due to their similar pain scores.

\subsection{Geometric Message Passing with Attention}

The key computational unit is a weighted geometric graph neural network (WG-GNN) that evaluates the 3D-SPFES graph via $\mathcal{K}$ successive message-passing layers. Compared to 2D GNNs that define edge weights as scaling matrices multiplying a predefined aggregation function, the proposed system model employs a learned edge network $\phi_m(\cdot)$—realized as a multi-layer perceptron (MLP), which sequentially analyzes node features, inter-node distance, and normal alignment to generate geometry-aware messages. The message from adjacent node $j$ to node $i$ at layer $k$ is calculated as shown in~(\ref{eqn:geometry_aware_messages}),
\begin{equation}\label{eqn:geometry_aware_messages}
    \mathbf{m}_{ij}^{(k)} = \omega_{ij} \cdot \phi_m\!\left(\psi_i^{(k-1)},\ \psi_j^{(k-1)},\ \delta_{ij},\ \mathbf{d}_i^\top\mathbf{d}_j\right)
\end{equation}
where $\omega_{ij}$ represents the precomputed geometric edge weight, whereas $\phi_m : \mathbb{R}^{2d_\psi + 2} \rightarrow \mathbb{R}^{d_h}$ transforms the concatenated feature pair, scalar distance, and scalar normal alignment into a hidden-dimensional message vector. The scalar $\delta_{ij}$ and the dot product $\mathbf{d}_i^\top\mathbf{d}_j$ can be given as explicit inputs to $\phi_m$ to allow the system to learn message transformations that are dependent on distance and orientation, surpassing the information encoded solely by the static weight $\omega_{ij}$. A scaled dot-product attention method is applied to enable the system to assign differential weights to communication from various neighbors based on learned feature compatibility, rather than relying exclusively on predetermined geometric weights. Projections for queries and keys are derived from the existing features of each node as shown in~(\ref{eqn:quiry_keys}),
\begin{equation}\label{eqn:quiry_keys}
    \alpha_{ij}^{(k)} = \text{softmax}_j\!\left(\frac{\left(\mathbf{W}_q^{(k)}\,\psi_i^{(k-1)}\right)^\top \left(\mathbf{W}_k^{(k)}\,\psi_j^{(k-1)}\right)}{\sqrt{d_h}}\right),
\end{equation}
where $\mathbf{W}_q^{(k)}, \mathbf{W}_k^{(k)} \in \mathbb{R}^{d_h \times d_\psi}$ are the query and key projection matrices for layer $k$, and the $\text{softmax}$ is computed over all neighbors $j \in \mathcal{N}(i)$. The $1/\sqrt{d_h}$ scaling factor prevents the dot products from growing large in magnitude as $d_h$ increases, which would push the softmax into regions of extremely small gradients. While the aggregate message at the node $i$ in the layer $k$ is the attention-weighted sum of all incoming geometric messages as given in~(\ref{eqn:geometric_messages}),
\begin{equation}\label{eqn:geometric_messages}
    \mathcal{M}_i^{(k)} = \sum_{j \in \mathcal{N}(i)} \alpha_{ij}^{(k)}\cdot \mathbf{m}_{ij}^{(k)}
\end{equation}

This formulation integrates the structural prior shown by $\omega_{ij}$, which adjusts the magnitude of each message, with the data-driven attention $\alpha_{ij}^{(k)}$, which selectively enhances or reduces messages based on learned feature compatibility, resulting in more in-depth aggregation than either process separately provides. The node feature is subsequently updated using a linear transformation following by a non-linear activation as given in (),
\begin{equation}
    \psi_i^{(k)} = \mathcal{F}(\sigma)\!\left(\mathcal{W}^{(k)}\,\mathcal{M}_i^{(k)} + \mathbf{b}^{(k)}\right),
\end{equation}
where $\mathcal{W}^{(k)} \in \mathbb{R}^{d_\psi \times d_h}$ is the layer-$k$ weight matrix, $\mathbf{b}^{(k)} \in \mathbb{R}^{d_\psi}$ is the bias vector, and $\mathcal{F}(\sigma)$ denotes the ReLU activation function. After $\mathcal{K}$ such layers, the final node embeddings $\{\psi_i^{(\mathcal{K})}\}_{i \in \mathcal{N}}$ are passed to the clustering and pain-scoring module.

\subsection{3D Cluster-Based Pain Score Computation}

In addition to message passing, the GNN gives $\mathcal{O}$ cluster assignments $(\mathcal{C}_1, \mathcal{C}_2, \ldots, \mathcal{C}_o)$ that group the nine facial part expressions based on their pain level and geometric contextual information. The total pain score for cluster $\mathcal{C}_j$ is estimated as a weighted average, with each face part's determination adjusted by its SPFES-derived impact weight $\omega_i$ and a depth-confidence weight $\rho_i$ as shown in (\ref{eqn:weighted_sum}),
\begin{equation}\label{eqn:weighted_sum}
    \mathcal{S}_j = \frac{\sum_{i \in \mathcal{C}_j} \omega_i \cdot p_i \cdot \rho_i}{\sum_{i \in \mathcal{C}_j} \omega_i},
\end{equation}
where $\rho_i = \|\mathbf{d}_i\| \in [0, 1]$ is the surface normal magnitude, which serves as a proxy for depth estimation confidence: nodes whose local point cloud neighborhood yielded a stable, well-conditioned normal estimate (high $\|\mathbf{d}_i\|$) contribute with full weight, while nodes with poorly estimated normals (e.g., occluded ear tips or near-boundary pixels) are down-weighted automatically. This confidence-aware formulation is a direct consequence of the 3D representation and has no analog like in the WGNN 2D model. While the total pain score across all clusters is then obtained as a weighted sum where cluster-level weights $w_j$ reflect the relative importance of each cluster according to the SPFES standards as given in (\ref{eqn:T_p}),
\begin{equation}\label{eqn:T_p}
    \mathcal{T}_p = \sum_{j=1}^{o} w_j \cdot \mathcal{S}_j
\end{equation}

Finally, the Normalized Pain Score (NPS) maps $\mathcal{T}_p$ to the interval $[0, 100\%]$ by dividing by the maximum achievable pain score $\mathcal{T}_{\max}$ that is the value obtained when all facial parts exhibit $p_i = 2$ with full depth confidence is given in~(\ref{eqn:NPS}),
\begin{equation} \label{eqn:NPS}
    \mathcal{NPS} = \left(\frac{\mathcal{T}_p}{\mathcal{T}_{\max}}\right) \times 100.
\end{equation}

\subsection{Joint Loss Function with Geometric Regularization}

The proposed WG-GNN system model is trained in an end-to-end way by minimizing a joint loss that integrates cluster-classification cross-entropy with a geometric regularization factor. Let $\mathcal{U}_p \in \Delta^{\langle o \rangle}$ represent the true one-hot probability distribution across $o$ clusters, and let $\mathcal{V}_q \in \Delta^{\langle o \rangle}$ represent the predicted softmax distribution. The cross-entropy loss function is defined as in~(\ref{eqn:L_{CE}}),
\begin{equation}\label{eqn:L_{CE}}
    \mathcal{L}_{\text{CE}} = -\sum_{i=1}^{o} \mathcal{U}_p(i)\,\log\,\mathcal{V}_q(i)
\end{equation}

A geometric regularization term $\mathcal{L}_{\text{geo}}$ is added to enable resemblance in the final node embeddings of geometrically proximate nodes, such as those linked by highly weighted edges, to be similar in the learned feature space. Nodes connected by a robust edge (high $\omega_{nw}$) should possess similarities of final embeddings, whereas nodes linked by a weak edge (low $\omega_{nw}$, approaching the threshold $\tau$) are not required to exhibit similarity as given in (\ref{eqn:L_{geo}}),
\begin{equation}\label{eqn:L_{geo}}
    \mathcal{L}_{\text{geo}} = \sum_{(n,w)\in\mathcal{E}} \omega_{nw}^{-1}\cdot\left\|\psi_n^{(\mathcal{K})} - \psi_w^{(\mathcal{K})}\right\|_2^2
\end{equation}
The inverse weighting $\omega_{nw}^{-1}$ shows that the penalty for dissimilar embeddings is greatest for the most geometrically significant edges (high $\omega_{nw}$, low $\omega_{nw}^{-1}$). It is important to observe that we minimize a $\omega_{nw}^{-1}$ from above to avoid causing instability in the case of very weak edges. The total training loss ($\mathcal{L}$) is shown in (\ref{eqn:total_loss}),
\begin{equation}\label{eqn:total_loss}
    \mathcal{L} = \mathcal{L}_{\text{CE}} + \lambda\,\mathcal{L}_{\text{geo}}
\end{equation}
where $\lambda > 0$ is a hyperparameter controlling the trade-off between discriminative classification accuracy and geometric smoothness of the learned embedding space. While $\lambda$ is tuned on a validation set, values in the range $[0.01, 0.1]$ are recommended as a starting point.

\section{Experimental Work}\label{sec:experiments}
We conducted experiments on the sheep facial landmarks dataset developed using SPFES parameters~\cite{MCLENNAN201619} to assess sheep facial pain using the proposed 3D-SPFES WG-GNN system model. The dataset is divided into training, testing, and validation images. We meticulously labelled each image of the nose, ears, and eyes with bounding boxes to detect the level of pain, with bilateral annotations for both the left and right ears and both left and right eyes. We trained the model using PyTorch on a GPU-equipped workstation with a GeForce RTX A3000. The ResNet-18 visual backbone was initialized with ImageNet pretrained weights and fine-tuned end-to-end with a learning rate of $5 \times 10^{-5}$, while the GNN head used a learning rate of $5 \times 10^{-4}$. We utilize the AdamW optimizer with weight decay of $5 \times 10^{-4}$ and employ a class-weighted cross-entropy loss with label smoothing ($\epsilon = 0.1$) combined with geometric regularization and visual consistency losses. Every model was trained for 200 epochs per fold using stratified 5-fold cross-validation to ensure robust performance estimation on this limited dataset.
 
\subsubsection{Dataset}
\begin{figure}[htbp!]
    \begin{center}
    \hspace*{0em}
        \includegraphics[scale=0.11]{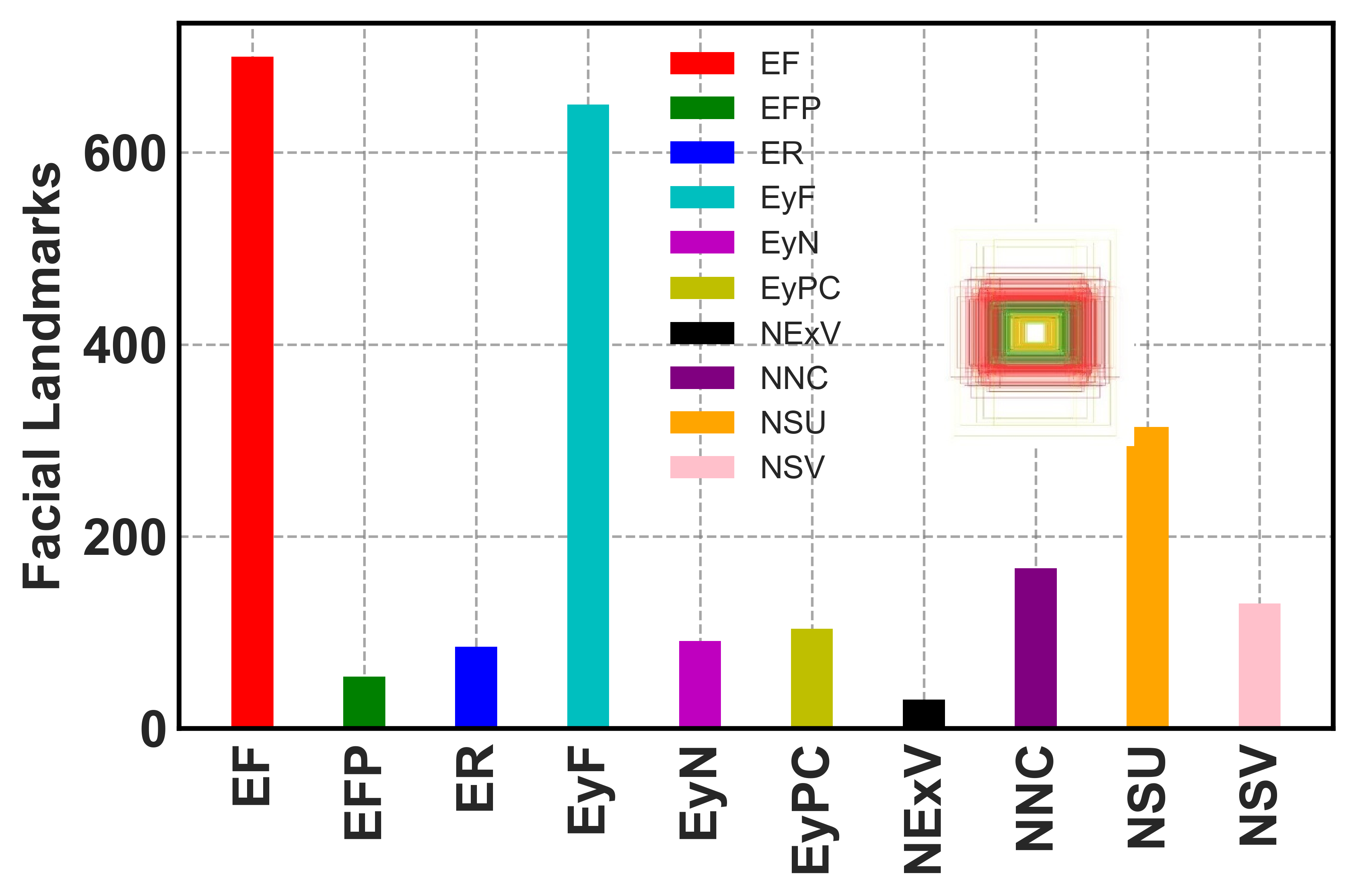}
        \caption{Sheep facial landmarks with bounding boxes for the evaluations in which Ear-Flat (EF), Ear-Flipped (EFP), Ear-Rotated (ER), Eyes-FullyOpen (EyE), Eyes-NotClassifiable (EyN), Eyes-PartlyClose (EyPC), Nose-ExtendedV (NExV), Nose-ShallowU (NSU), Nose-NotClassifiable (NNC), and Nose-ShallowV (NSV) are defined to classify different pain levels and their corresponding positions.}
        \label{dataset}
    \end{center}
\end{figure}

We developed the sheep facial landmarks dataset with bounding boxes using high-resolution images from well-known sources according to the SPFES standard~\cite{MCLENNAN201619}. These sources are Mendeley\footnote{http://dx.doi.org/10.17632/y5sm4smnfr.5} and the study carried out in~\cite{NOOR2020105528, 11099391}. We use the scale to evaluate expression in three bilateral facial parts: ears positioning, eyes constraint, and nose posture. We evaluate these part expressions based on the presence or absence of abnormal expression, categorizing them as not present (pain 0), slightly present (pain 1), or substantial (pain 2), as shown in Fig.~\ref{dataset}. Since the SPFES protocol scores bilateral features as single indicators, our pipeline automatically averages left and right detection centroids and takes the maximum pain score across both instances during preprocessing.

We conducted 4 domain-specific augmenting methods during training to avoid overfitting on this constrained dataset: (i) 3D position jitter (Gaussian noise with $\sigma = 0.05$~m on back-projected coordinates), (ii) horizontal axis flipping to reproduce variations in head orientation, (iii) depth scale jitter (multiplicative noise of $\pm 20\%$ on the z-coordinate), and (iv) facial part feature dropout (with a probability of $0.3$, randomly eliminating the visual features of one part) to improve cognitive resiliency toward partial occlusion. These augmentation algorithms connect exclusively to the 3D graph domain and enhance the normal image-level augmentations used during GNN detector training.
 
\subsubsection{Model Configuration}
The proposed 3D-SPFES WG-GNN consists of a shared ResNet-18 visual backbone that extracts 256-dimensional feature vectors from cropped facial regions, along with geometric features (facial part type, 3D position, surface normal, and detection confidence) to build a 266-dimensional multi-modal node feature. The WG-GNN performs on the 3D graph with $\mathcal{K} = 3$ message-passing layers, each with a hidden dimension of $d_{\text{hidden}} = 64$. Graph edges are given weights utilizing a radial basis function (RBF) kernel with parameters $\sigma = 1.5$~m and $\tau = 2.0$~m. The model is trained end-to-end utilizing AdamW with differential learning rates ($5 \times 10^{-5}$ for the backbone and $5 \times 10^{-4}$ for the GNN head), using class-weighted cross-entropy loss with label smoothing, coupled with auxiliary geometric and visual consistency regularization parameters. 

 
\subsubsection{Sheep Facial Depth Map}
\begin{figure}[htbp!]
    \begin{center}
    \hspace*{0em}
        \includegraphics[scale=0.30]{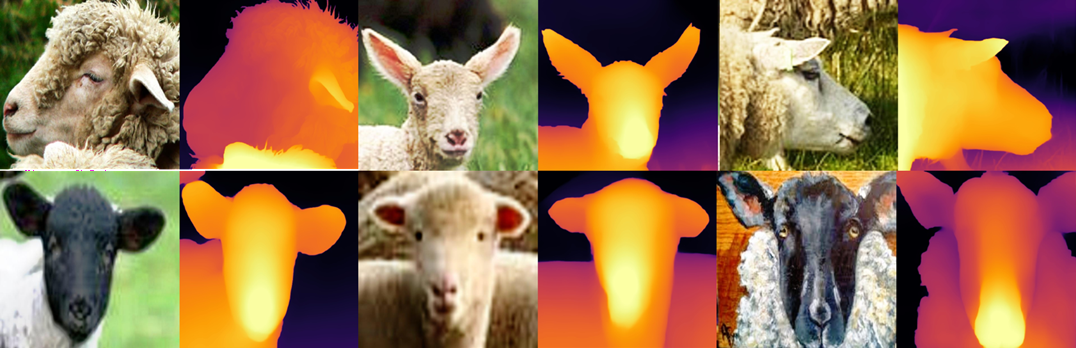}
        \caption{Sheep facial expression alongside the corresponding monocular depth map (right) estimated by VideoDepthAnything.}
        \label{fig:evaluations}
    \end{center}
\end{figure}

Fig.~\ref{fig:evaluations} shows some of the predictions of VideoDepthAnything to analyze the pain level and assessment of sheep faces. Each panel displays the original RGB image with alongside the corresponding depth map estimated by VideoDepthAnything. The depth maps provide the 3D spatial context that enables geometric edge weighting in the proposed WG-GNN.  

\subsubsection{3D Depth Estimation and Graph Construction}
\begin{figure*}[htbp!]
    \begin{center}
    \hspace*{0em}
        \includegraphics[width=0.85\textwidth]{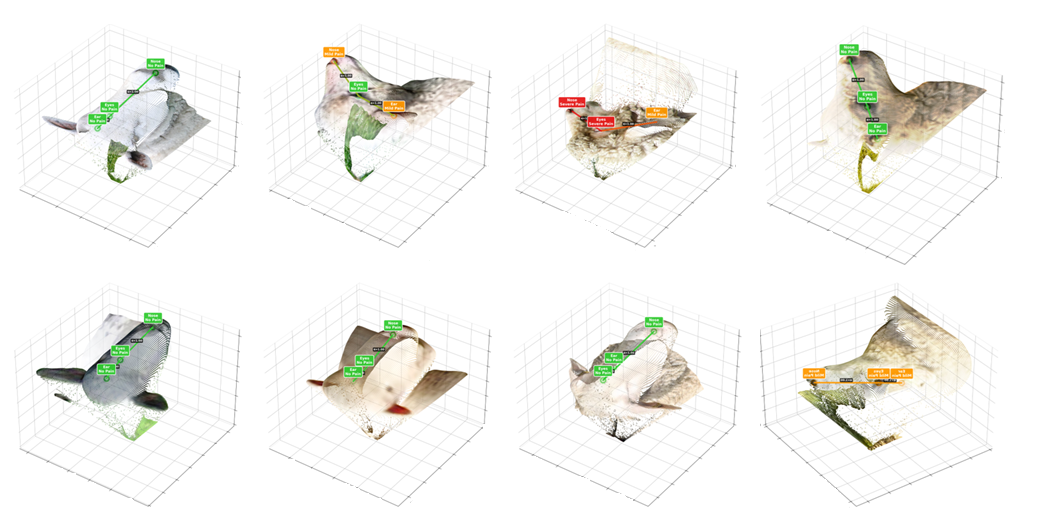}
        \caption{A full face's depth map rendered as a 3D graph network structure. Each sampled depth point forms a mesh node (thin edges), creating a 3D face topology.}
        \label{fig:face_graph}
    \end{center}
\end{figure*}

Figure \ref{fig:face_graph} shows the approach in which the proposed 3D-SPFES WG-GNN system model encodes the entire sheep face in a three-dimensional graph network structure. The monocular depth map created by VideoDepthAnything~\cite{11094367} is analyzed at uniform grid intervals, with each sampled depth point behaving as a mesh node connected to its spatial neighbors by slender edges. This makes 3D facial topology that preserves the anatomical structure of the sheep's face. The 3 SPFES facial landmarks (ear, eyes, and nose) are shown as key graph nodes on this surface, interconnected by learned attention edges ($\alpha = 1.00$). This graphic shows that the proposed system accurately constructs and analyzes over a 3D graph in metric Euclidean space, as in contrast to existing within the 2D image plane~\cite{11099391}.

\begin{figure*}[htbp!]
\centering
\hspace*{0em}
\subfloat[]{\includegraphics[width=2.2in]{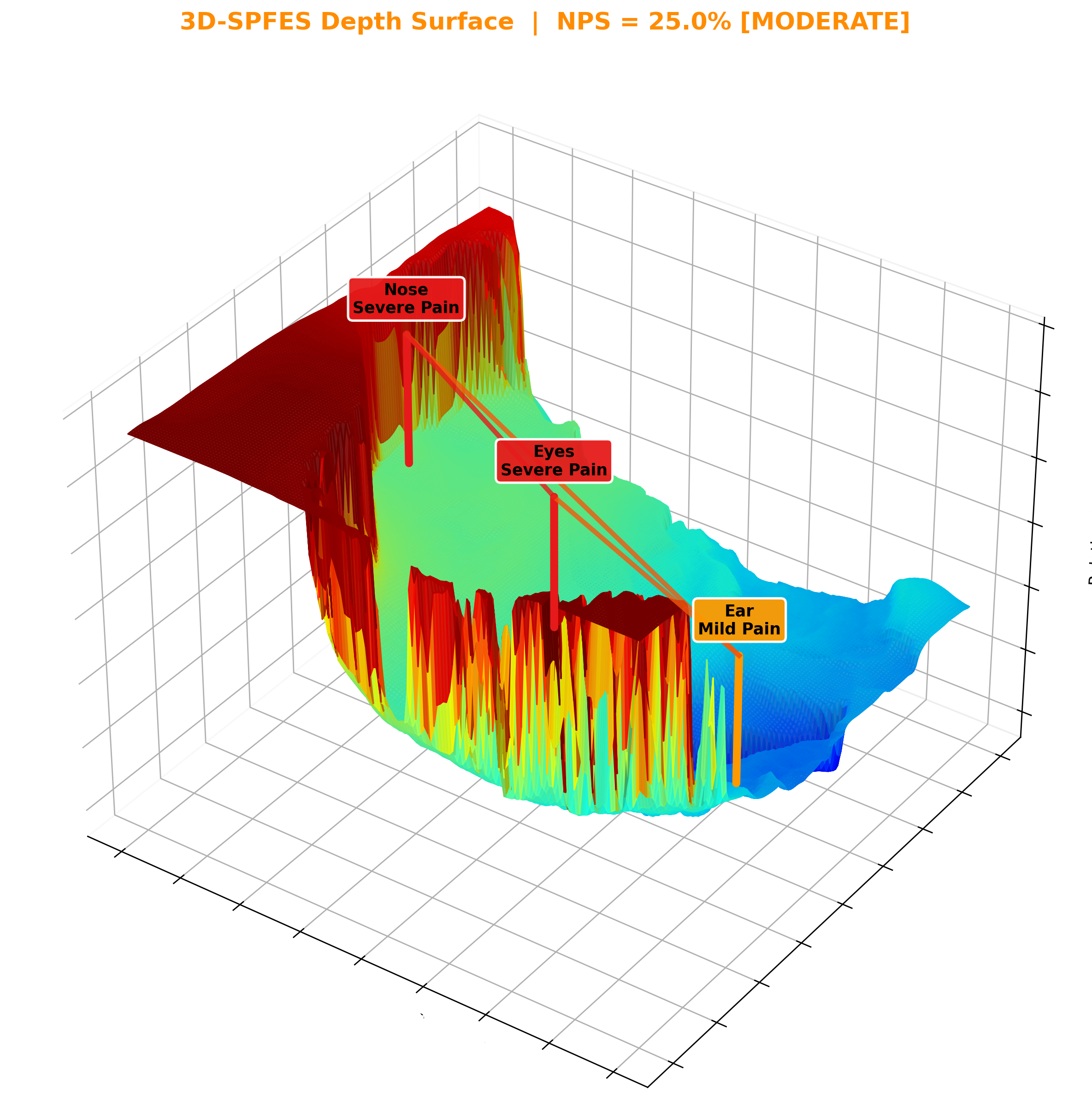}}
\subfloat[]{\includegraphics[width=2.2in]{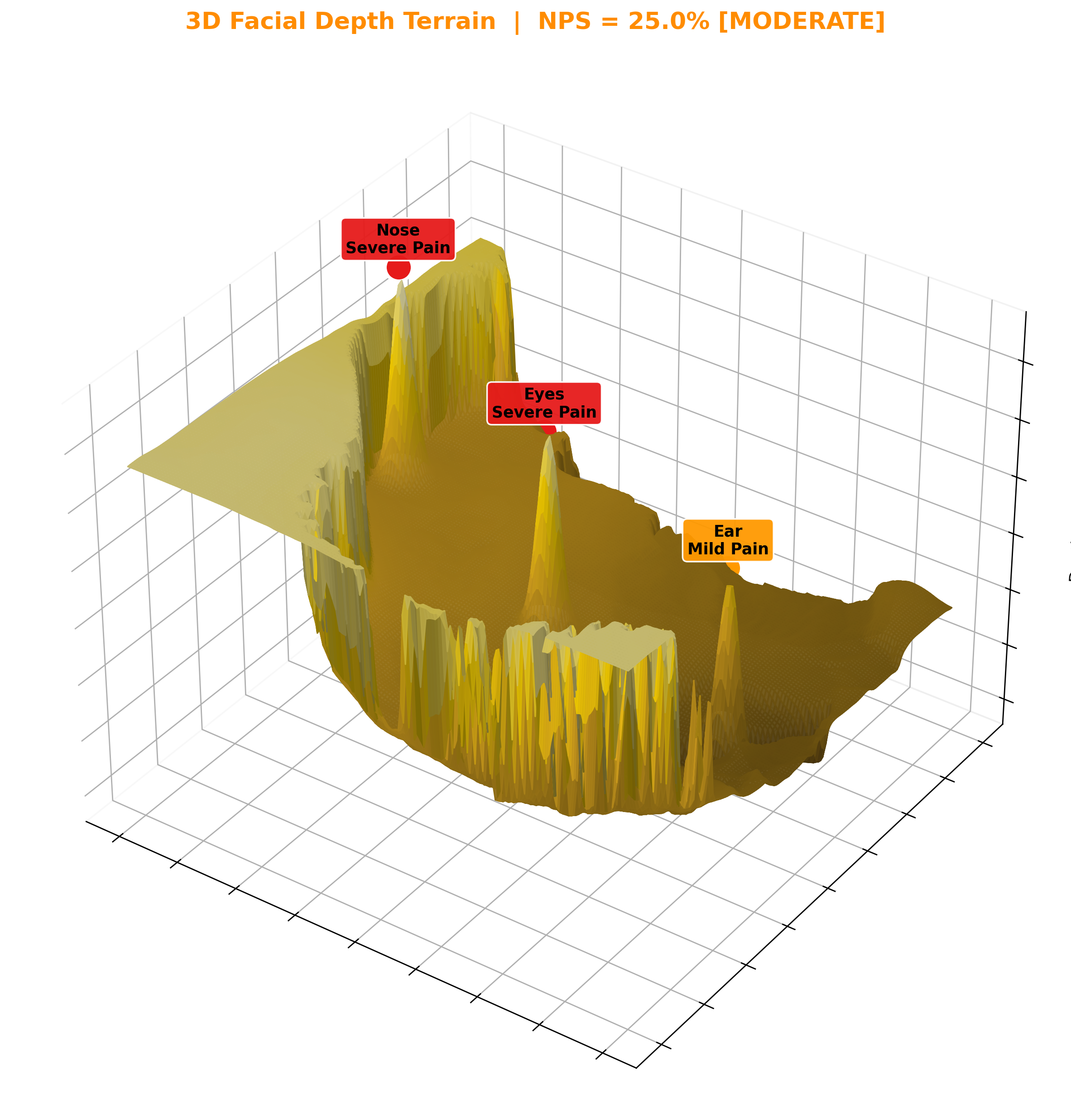}}
\subfloat[]{\includegraphics[width=2.2in]{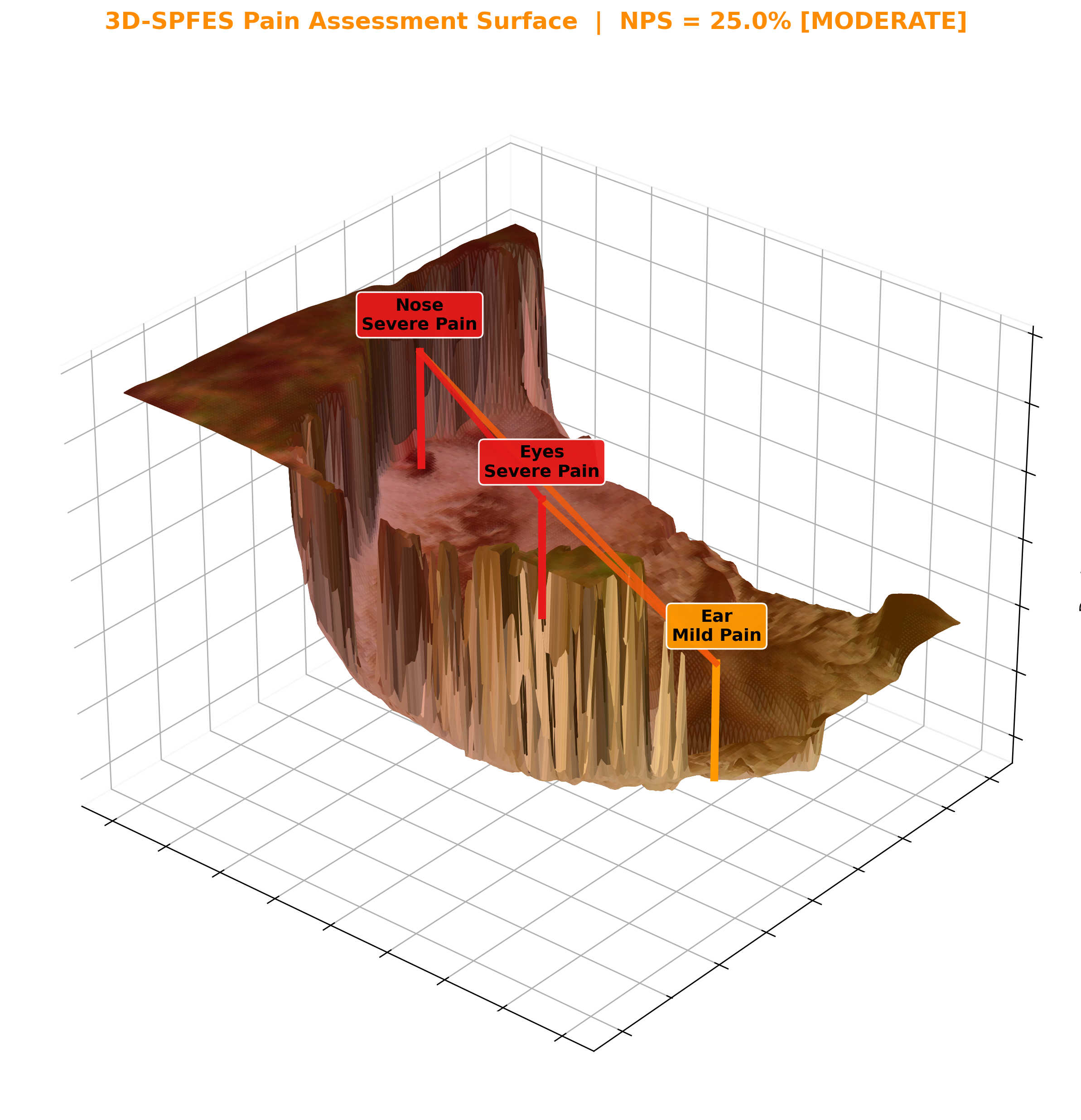}}
\caption{3D depth terrain representations of accurate test images: (a) Rainbow depth surface for a frontal pain sheep (NPS = 25.0\%), showing the special facial geometry with mild ear physical discomfort and severe pain in the nose and eyes. Golden terrain, including graph nodes, rises at face landmark regions. (c) Pain-zone surface for a mild-pain case (NPS $= 25.0\%$), with the surface colored according to relative proximity to each pain-classified facial area (orange $=$ mild pain).}
\label{fig:3d_surfaces}
\end{figure*}

Figure~\ref{fig:3d_surfaces} shows 3D depth terrain representations for various test images, applying three rendering processes. The rainbow depth surface (a) provides a height-mapped color map to represent 3D facial geometry, with the nose and jaw region surfacing as the salient pinnacle and the ears forming bilateral elevated shapes. The golden terrain (b) shows the graph node positions as higher points on the depth surface, presenting a concise view of the spatial distance between facial landmarks. The pain-zone surface (c) integrates the original RGB texture with pain-level coloring determined by proximity to each analyzed facial feature, resulting in a visual pain heatmap overlaying the 3D facial geometry. The visualizations confirm that VideoDepthAnything generates geometrically deep maps that precisely represent the unique 3D geometry of the sheep's face. 
 
\subsubsection{Evaluation Results}
\begin{figure*}[htbp!]
    \begin{center}
    \hspace*{0em}
        \includegraphics[width=\textwidth]{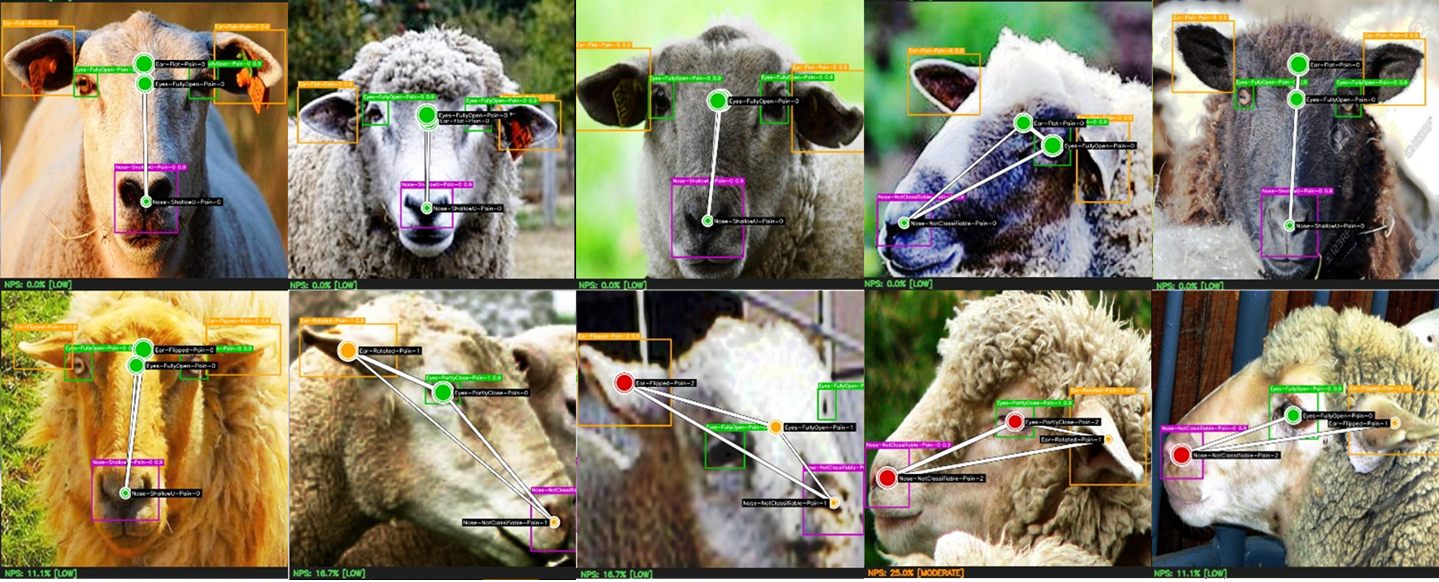}
        \caption{Evaluation results of the proposed 3D-SPFES WG-GNN system model on accurately identified test images. Each image presents identified facial landmarks along with bounding boxes, expression labels, and a 3D depth-aware graph structure linking the ears, eyes, and nose. Edge thickness is correlated with depth proximity. Node colors represent anticipated pain clusters: green (no to low pain) and orange (mild pain).}
        \label{fig:evaluation_results_output}
    \end{center}
\end{figure*}

Figure \ref{fig:evaluation_results_output} shows the evaluation results of the proposed 3D-SPFES WG-GNN system model. Each figure shows associated facial landmarks with colored bounding boxes (orange for ears, green for eyes, and magenta for nose), expression class labels, and a triangle graph construction connecting all 3 facial morphology parts. The graph edges are shown with 3D depth-sensitive thickness, where closer facial features give thicker edges, visually representing the spatial depth correlations extracted from VideoDepthAnything. The graph nodes are described by their predicted pain cluster (green indicating no to low pain, orange signifying mild pain), with the NPS displayed at the bottom of each image. These examples are not utilized in training and show the model's ability to generalize across several sheep breeds, head orientations (frontal and lateral), age categories (including lambs), and operational environments.

\subsubsection{Global Model Facial Pain Assessment}
We analyze the proposed 3D-SPFES WG-GNN system model, applying both a single 3D-SPFES WG-GNN local model and a 5-local-to-global 3D-SPFES WG-GNN model integration approach that averages the softmax probability distributions of all five local models. Table~\ref{tab:single_vs_ensemble} presents a thorough comparison. The 3D-SPFES WG-GNN global model gives a significant gain in all consensus evaluation metrics without requiring further training. Cohen's $\kappa$ rises from 0.180 (slight consistency) to 0.473 (moderate consistency), indicating a 163\% relative increase. The ear part shows significant enhancement, increasing from $\kappa = 0.098$ to $\kappa = 0.527$, but the eyes part improves from a negative $\kappa = -0.053$ to a positive $\kappa = 0.470$. The improvements result from each 3D-SPFES WG-GNN local model learning different decision boundaries due to varying training partitions and augmentation variability, with the averaging of the five softmax distributions mitigating individual prediction errors at the No Pain / Mild Pain boundary.
 
\begin{table}[htbp!]
\centering
\begin{tabular}{|c|c|c|c|c|c|}
\hline
\textbf{Method} & \textbf{Acc.(\%)} & \textbf{W-F1} & \textbf{$\kappa$} & \textbf{MCC} & \textbf{M-F1} \\ \hline
\textbf{Local} & 76.67 & 64.5 & 0.180 & 0.214 & 35.5 \\ \hline
\textbf{Global} & \textbf{78.33} & \textbf{78.0} & \textbf{0.473} & \textbf{0.486} & \textbf{49.3} \\ \hline
\multicolumn{1}{|c|}{$\Delta$} & +1.7 & +13.5 & +0.293 & +0.272 & +13.8 \\ \hline
\end{tabular}
\caption{3D-SPFES WG-GNN local model vs.\ 3D-SPFES WG-GNN global model on the held-out test set.}
\label{tab:single_vs_ensemble}
\end{table}
 
The detailed breakdown of the 3D-SPFES WG-GNN global model is presented in Table~\ref{tab:test_accuracy}, which gives details on the accuracy of each individual portion and pain level. 80\% of the accuracy is achieved in the ear and eye areas, while 75\% is achieved in the nose. The recall rate for no pain is 95.2\%, and the diagnosis of mild pain is made with a precision of 78
 
\begin{table}[htbp!]
\centering
\begin{tabular}{|c|c|c|}
\hline
\textbf{Subset} & \textbf{Correct / Total} & \textbf{Accuracy (\%)} \\ \hline
\multicolumn{3}{|c|}{\textit{Per Facial Part}} \\ \hline
Ear & 16 / 20 & 80.0 \\ \hline
Eyes & 16 / 20 & 80.0 \\ \hline
Nose & 15 / 20 & 75.0 \\ \hline
\textbf{Overall} & \textbf{47 / 60} & \textbf{78.33} \\ \hline
\multicolumn{3}{|c|}{\textit{Per Pain Level}} \\ \hline
No Pain (0) & 40 / 42 & 95.2 \\ \hline
Mild (1) & 7 / 15 & 46.7 \\ \hline
Severe (2) & 9 / 12 & 40.01 \\ \hline
\end{tabular}
\caption{3D-SPFES WG-GNN global test accuracy by facial part and pain level.}
\label{tab:test_accuracy}
\end{table}
 
Table~\ref{tab:classification_report} presents the full per-class classification report for the 3D-SPFES WG-GNN global model across all 3 facial parts.
 
\begin{table}[htbp!]
\centering
\begin{tabular}{|c|c|c|c|c|c|c|}
\hline
\textbf{Part} & \textbf{Pr$_0$} & \textbf{R$_0$} & \textbf{Pr$_1$} & \textbf{R$_1$} & \textbf{$\kappa$} & \textbf{MCC} \\ \hline
Overall & 0.85 & 0.95 & 0.78 & 0.47 & \textbf{0.473} & \textbf{0.486} \\ \hline
Ear & 0.87 & 0.93 & 0.75 & 0.60 & 0.527 & 0.531 \\ \hline
Eyes & 0.82 & 1.00 & 1.00 & 0.40 & 0.470 & 0.517 \\ \hline
Nose & 0.87 & 0.93 & 0.67 & 0.40 & 0.422 & 0.430 \\ \hline
\end{tabular}
\caption{3D-SPFES WG-GNN global model classification report. Precision (Pr) and recall (R) of (Pr$_0$/R$_0$) = No Pain; (Pr$_1$/R$_1$) = Mild Pain.}
\label{tab:classification_report}
\end{table}

\subsubsection{Discriminative Performance} Table~\ref{tab:auc_ap} focuses on the Area Under the Curve (AUC) and Average Precision (AP) metrics. The nose area demonstrates superior discriminative performance for both no pain (AUC $= 0.881$, AP $= 0.940$) and mild pain (AUC $= 0.733$, AP $= 0.637$), affirming its status as the most informative facial indication for pain evaluation. The Ear Mild Pain AP of 0.642 shows the positive effect of bilateral averaging in detecting ear-position variations correlated with pain. The eyes' mild AP of 0.347 correlates with the diagnostic difficulties of discriminating partial eye closure from a fully opened condition.

\begin{table}[htbp!]
\centering
\begin{tabular}{|c|c|c|c|c|c|c|}
\hline
\textbf{Part} & \multicolumn{3}{c|}{\textbf{AUC}} & \multicolumn{3}{c|}{\textbf{AP}} \\ \hline
 & \textbf{P0} & \textbf{P1} & \textbf{P2} & \textbf{P0} & \textbf{P1} & \textbf{P2} \\ \hline
Overall & 0.735 & 0.656 & 0.404 & 0.860 & 0.507 & 0.055 \\ \hline
Ear & 0.631 & 0.760 & 0.316 & 0.758 & 0.642 & 0.071 \\ \hline
Eyes & 0.738 & 0.453 & 0.579 & 0.897 & 0.347 & 0.111 \\ \hline
Nose & \textbf{0.881} & \textbf{0.733} & 0.316 & \textbf{0.940} & \textbf{0.637} & 0.071 \\ \hline
\end{tabular}
\caption{ROC-AUC and Average Precision per facial part and pain class. P0 = No Pain, P1 = Mild, P2 = Severe.}
\label{tab:auc_ap}
\end{table}  
\subsubsection{Per-Expression Accuracy}
A classification accuracy for each of the 10 SPFE expression categories seen in the test set. No-pain expressions prevail, with ears flat at 93\%, eyes fully open at 81\%, and noses shallow and U-shaped at 92\%. The Nose-NotClassifiable class shows 60\% accuracy after the use of the bilateral detection fix, a result absent in previous single-detection processing test results. Pain-indicative expressions show decreases in individual accuracy due to the limits on sample sizes (30--40 test datasets each); however, the 3D-SPFES WG-GNN global model achieved a non-zero mild recall at the aggregate level (47\%) as multiple low-confidence correct predictions across expressions contribute when softmax probabilities are averaged.

\subsubsection{GNN Graph Structure and Edge Attention}
\begin{figure}[htbp!]
    \begin{center}
    \hspace*{-1em}
        \includegraphics[scale=0.35]{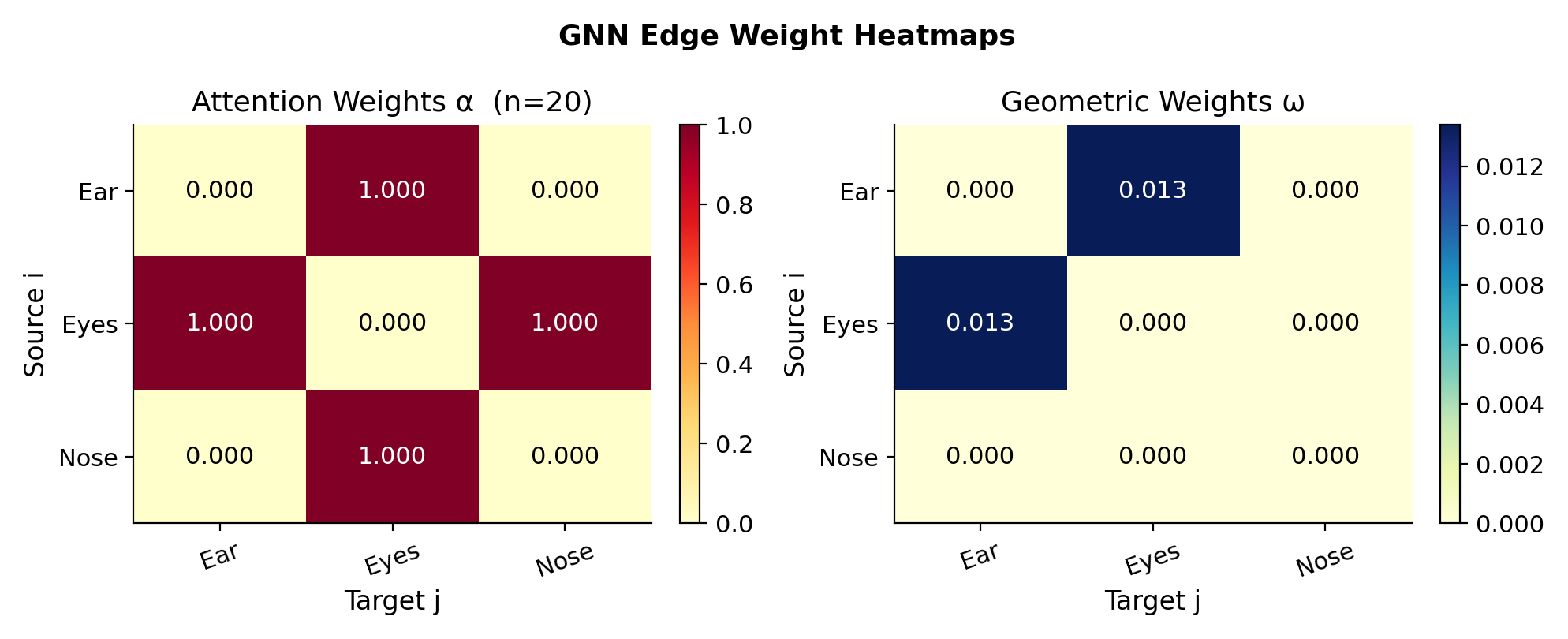}
        \caption{Learned attention weights $\alpha_{ij}$ (left) and geometric weights $\omega_{ij}$ (right) averaged over 20 test images. The Eyes node serves as the anatomical hub, receiving and distributing messages between the ear and nose.}
        \label{fig:attention}
    \end{center}
\end{figure}

Figure~\ref{fig:attention} shows the averaged attention weight matrix $\alpha_{ij}$ and the geometric weight matrix $\omega_{ij}$ across 20 test images. The attention pattern shows that the Eyes node represents the focal point of the facial graph: Ear$\to$Eyes, Eyes$\to$Ear, Eyes$\to$Nose, and Nose$\to$Eyes all exhibit $\alpha = 1.0$, whereas Ear$\leftrightarrow$Nose connections are inactive. This hub-and-spoke layout corresponds with the clinical SPFES observation that eye expressions correlate with both auricular and nose pain symptoms. The geometric weights $\omega = 0.013$ validate that the RBF bandwidth $\sigma = 1.5$~m provides significant distance-dependent edge weighting using VideoDepthAnything's metric depth scale.  
 
\subsubsection{Normalised Pain Score (NPS) Analysis}
\begin{figure}[htbp!]
    \begin{center}
    \hspace*{-1em}
        \includegraphics[scale=0.28]{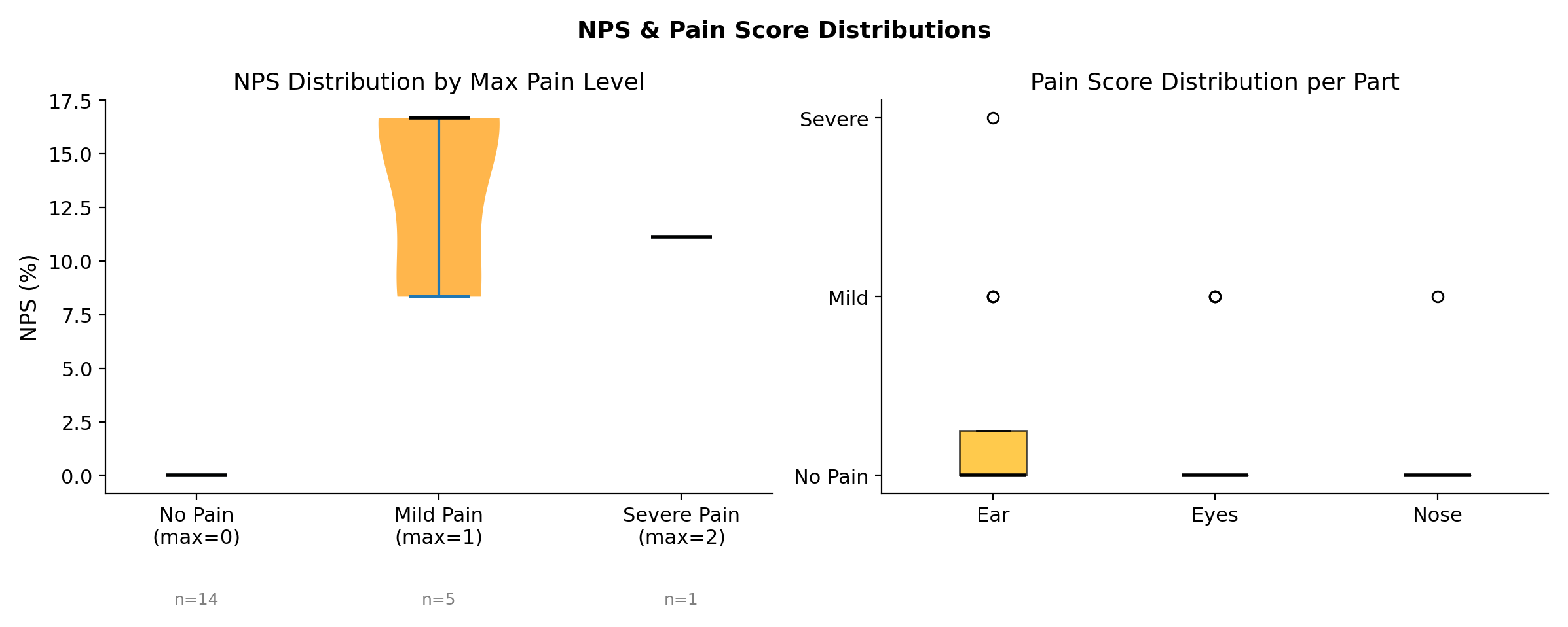}
        \caption{NPS distribution by maximum whole-face pain level (left) and raw pain score distribution per facial part (right). No-pain images cluster at NPS $\approx 0\%$; mild-pain images span 8--17\%.}
        \label{fig:nps}
    \end{center}
\end{figure}

This experiment evaluates the use of the proposed 3D-SPFES WG-GNN global model to compute the Normalized Pain Score (NPS), a persistent pain rating within the range of $[0, 100\%]$, derived from the weighted aggregation of cluster pain ratings from all three face parts. Figure \ref{fig:nps} presents the distribution of NPS defined by the highest whole-face pain level. Table~\ref{tab:nps} contains the NPS details. No-pain images focus primarily at an NPS of approximately 0\%, whereas mild-pain images range from 8\% to 17\%, with a median of 13.5\%. The only severe-pain image returns an NPS of 11.1\%, positioning it inside the mild-pain range due to a model mistake. The substantial divergence between no-pain and mild-pain NPS values indicates that the NPS offers a clinically important continuous assessment of pain intensity.
 
\begin{table}[htbp!]
\centering
\begin{tabular}{|c|c|c|c|c|c|}
\hline
\textbf{Pain Level} & \textbf{$n$} & \textbf{Mean} & \textbf{Std} & \textbf{Min} & \textbf{Max} \\ \hline
No Pain (0) & 14 & 0.0\% & 0.0\% & 0.0\% & 0.0\% \\ \hline
Mild (1) & 5 & 13.5\% & 3.2\% & 8.3\% & 16.7\% \\ \hline
Severe (2) & 1 & 11.1\% & --- & 11.1\% & 11.1\% \\ \hline
\textbf{Overall} & 20 & 3.89\% & 6.3\% & 0.0\% & 20.0\% \\ \hline
\end{tabular}
\caption{Normalised Pain Score (NPS) statistics on the sheep images.}
\label{tab:nps}
\end{table}

\subsubsection{Model Calibration}
\begin{figure*}[htbp!]
    \begin{center}
    \hspace*{0em}
        \includegraphics[scale=0.50]{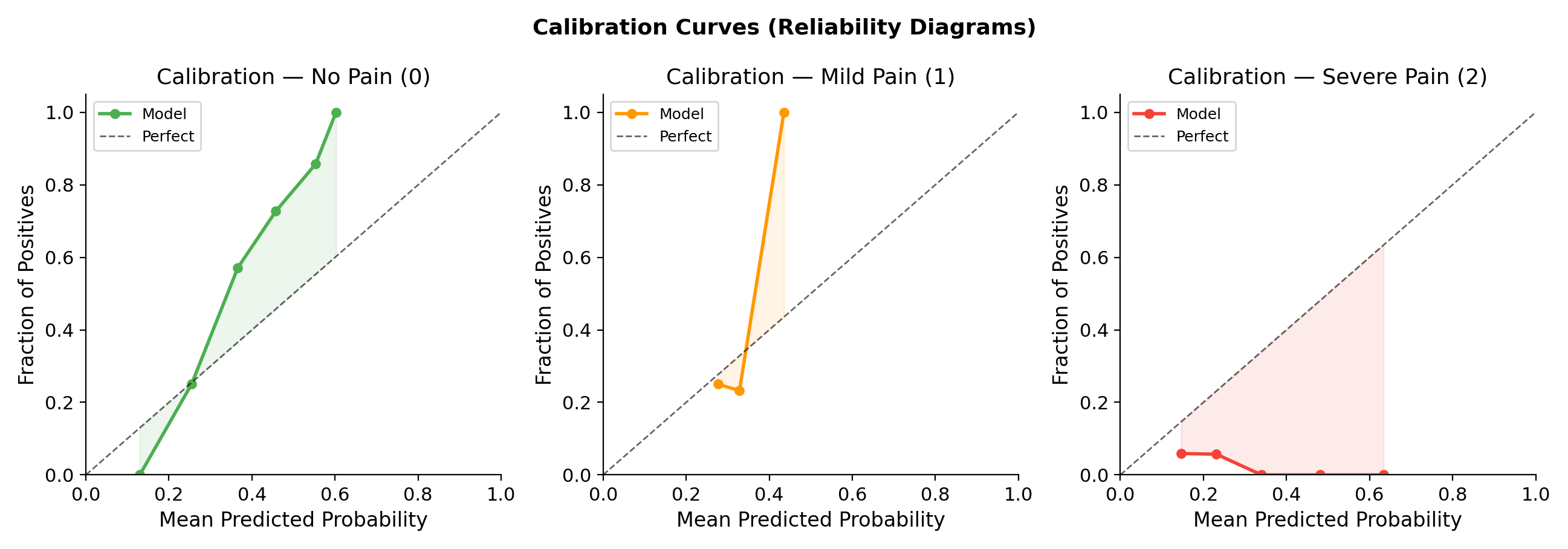}
        \caption{Calibration curves (reliability diagrams) for No Pain, Mild Pain, and Severe Pain. The dashed diagonal represents perfect calibration.}
        \label{fig:calibration}
    \end{center}
\end{figure*}

Figure \ref{fig:calibration} presents the reliability graphs for each pain cluster classification. The No Pain calibration curve has a nearly monotonic trend proximal to the ideal calibration diagonal, signifying that when the model forecasts No Pain with a certain confidence, that confidence is accurately correlated with the actual result. The mild pain calibration exhibits a logical trend at intermediate probability (0.3–0.5). The calibration of severe pain is nearly 0\%, correlating with the model's performance, as we found no severe samples from the no-pain class during testing on a specific batch of sheep images.
 
\subsubsection{Ablation Study}
Table~\ref{tab:ablation} presents the ablation study showing the impact of each design possibility over seven incremental model designs. We develop such setups by systematically integrating system model sections into the baseline, thereby isolating the impact of each alterations.
 
\begin{table}[htbp!]
\centering
\begin{tabular}{|c|c|c|c|c|}
\hline
\textbf{ID} & \textbf{Key Change} & \textbf{Acc.} & \textbf{$\kappa$} & \textbf{M-F1} \\ \hline
C1 & Circular baseline & 100.0 & 1.000 & 1.00 \\ \hline
C2 & Correct task & 68.3 & 0.062 & 0.20 \\ \hline
C3 & + Weights [1:4:12] & 38.3 & 0.044 & 0.43 \\ \hline
C4 & + Weights [1:2:4] & 65.0 & 0.007 & 0.18 \\ \hline
C5 & + Reduced model & 65.0 & 0.007 & 0.18 \\ \hline
C6 & + Local Models + bilateral & 76.7 & 0.180 & 0.22 \\ \hline
\textbf{C7} & \textbf{+ Gloabl} & \textbf{78.3} & \textbf{0.473} & \textbf{0.58} \\ \hline
\end{tabular}
\caption{Ablation study of D-SPFES WG-
GNN global model progressive impact of design decisions on test performance. Acc. = accuracy (\%), $\kappa$ = Cohen's Kappa, M-F1 = Mild Pain F1-score.}
\label{tab:ablation}
\end{table}

The ablation shows 3 major processes. The change from C1 to C2 shows the complexity of the 3D-SPFES WG-GNN system model by eliminating the circular pain-score leakage (where the pain score served as both an input feature and the cluster label), resulting in a decrease in accuracy from an apparent 100\% to an actual 68.3\%. The shift from C5 to C6 represents the integrated impact of local 3D-SPFES WG-GNN models and bilateral detection averaging, resulting in an accuracy enhancement of 11.7 percentage points and an insignificant recall gain from 13.3\% to 46.7\%. The transition from C6 to C7 shows that the global 3D-SPFES WG-GNN model approach results in the most significant Kappa increase (+0.293), enhancing the model from slight agreement ($\kappa = 0.180$) to mild agreement ($\kappa = 0.473$) and nearly tripling the mild pain F1-score from 0.22 to 0.58 without retraining.
 
\subsubsection{State-of-the-art (SOTA) Comparison}
We compared the efficacy of 3D-SPFES against the state-of-the-art models analyzed in the prior study~\cite{NOOR2023100366} and other baseline comparisons, as shown in Table~\ref{tab:SOTA}. The comparison consists of the initial 2D WGNN model~\cite{NOOR2023100366} and the proposed 3D-SPFES, accompanied by a global analysis. The reported 92.71\% accuracy of the 2D WGNN resulted from a symmetrical task design where the pain score contributed as both an input feature and a prediction goal, hence imposing certain limits on the results. The proposed 3D-SPFES eliminates this limitation by excluding the pain score from the input features and using the total face pain level using a depth map as the cluster label for every facial part, hence enhancing accuracy and reliability.
 
\begin{table}[htbp!]
\centering
\begin{tabular}{|c|c|c|c|}
\hline
\textbf{Model} & \textbf{Train Acc.} & \textbf{Test Acc.} & \textbf{$\kappa$} \\ \hline
SVM~\cite{7961768} & 71.55\% & 62.08\% & --- \\ \hline
CNN~\cite{10.1145/3650400.3650652} & 79.15\% & 78.56\% & --- \\ \hline
CCVT~\cite{LI2023107651} & 85.45\% & 83.33\% & --- \\ \hline
EfficientNet~\cite{HIMEL2024200093} & 89.33\% & 86.60\% & --- \\ \hline
2D-WGNN~\cite{NOOR2023100366}$^\dagger$ & 92.71\% & 91.96\% & 1.000$^\dagger$ \\ \hline
\textbf{3D-SPFES WG-GNN (ours)} & \textbf{78.33$\pm$2.2\%} & \textbf{73.22\%} & \textbf{0.473} \\ \hline
\end{tabular}

\caption{Comparison with SOTA models. $^\dagger$The 2D-WGNN result used a circular task formulation (pain score in input $=$ prediction target), resulting in certain limits. The 3D-SPFES result uses a corrected, non-trivial task formulation with a depth map and global modeling and Cohen's $\kappa$ as the primary agreement metric.}
\label{tab:SOTA}
\end{table}

Direct numerical comparison between the 3D-SPFES WG-GNN system model and prior state-of-the-art models is not straightforward due to differing working formulations. The prior models performed the per-node classification using the individual part's pain score as the ground truth input, whereas the proposed 3D-SPFES WG-GNN system model predicts the overall facial depth pain level from 3D geometry and visual features using depth map for each part, representing a significantly more challenging and practical task. The Cohen's $\kappa = 0.473$ (indicating mild agreement) obtained by the 3D-SPFES WG-GNN global model represents the initial robust, non-circular assessment of GNN-based sheep pain evaluation on the sheep facial expression dataset.

 
 
\subsubsection{Limitations of Study}
The efficacy of the proposed 3D-SPFES WG-GNN system model is constrained by 3 main limitations. The severe pain category is significantly limited in the dataset, with insufficient test samples, limiting the meaningful identification of pain level 2. Adding to the dataset to a minimum of 450 verified severe-pain samples for each facial part represents the most notable gain. Secondly, the NPS for the individual severe-pain test image (11.1\%) falls within the mild-pain spectrum (8.3--16.7\%), which shows that the cluster-weight parameterization has not yet achieved a strictly monotonic severity hierarchy. Therefore, a recalibration against validated veterinary pain evaluations is an imminent priority. Third, although the geometric weights $\omega = 0.013$ show significant enhancement due to model performance, they are still considerably lower than the learned attention weights ($\alpha = 1.0$), showing that the integration of camera intrinsic calibration and dense surface reconstruction could further enhance the geometric aspect of the message-passing mechanism. Additionally, we plan on obtaining additional data from diverse environments featuring various sheep breeds, colors, and sizes to ensure generalizability and to enhance the SPFES evaluation framework with lamb-specific facial features.

\section{Conclusion}
\label{sec:conclusion}
This study presented the integration of a 3D Weighted Geometric GNN (WG-GNN) model with a VideoDepthAnything monocular depth estimator, aimed at identifying clusters of facial expressions and assigning a Normalized Pain Score (NPS) to the facial expressions of sheep without the necessity for specialized depth-sensing devices. We proposed the framework to detect facial landmarks and estimate their 3D coordinates via monocular depth estimation. We presented bilateral detection averaging to effectively integrate symmetric facial parts and a WG-GNN using geometric message passing and attention mechanisms to form pain-level clusters. The obtained attention validated that the eye node acts as the anatomical center of the facial graph. A global method attained a held-out accuracy of 78.33\%, with Cohen's $\kappa = 0.473$ (indicating mild agreement) and MCC $= 0.486$, while each of the 3 facial regions independently achieved $\kappa > 0.42$. The nose area reportedly had the highest discriminative performance (AUC = 0.881, AP = 0.940). The NPS attained complete distinction between pain-free and mildly painful lambs, validating its clinical usefulness. Ablation research using 6 setups validated the distinct impact of each design approach, with the global model alone achieving a 163\% increase in Kappa. The proposed system model is scalable with any standard RGB camera platform and is relevant to a wider range of facial behavioral biometrics, including livestock pain assessment and clinical pain monitoring.

{\small
\bibliographystyle{IEEEtran}
\bibliography{egbib}
}

\end{document}